%% file: paper.tex
\documentclass{article} % For LaTeX2e
\usepackage{iclr2021_conference,times}

\input{math_commands.tex}
\newcommand{\norm}[1]{\left\lVert#1\right\rVert}
\usepackage{bbm}
\usepackage{amsthm}
\newtheorem{theorem}{Theorem}
\usepackage[ruled,vlined ]{algorithm2e}
\usepackage{hyperref}
\usepackage{url}
\usepackage{graphicx}
\usepackage{subfig}
\usepackage{float}
\usepackage{titlesec}
\usepackage{stmaryrd}

\title{Learning Fair Canonical Polyadical Decompositions using a Kernel Independence Criterion}

\author{Kevin Kim, Alex Gittens  \\
Department of Computer Science, Rensselaer Polytechnic Institute \\

\texttt{\{kimk15, gittea\}@rpi.edu}
}

\iclrfinalcopy % Uncomment for camera-ready version, but NOT for submission.
\titlespacing*{\section}{0pt}{1.5ex}{1.5ex}
\titlespacing*{\subsection}{0pt}{1.5ex}{1.5ex}
\begin{document}

\maketitle
\begin{abstract}
%Provided enough data and computational power, many classification and regression tasks in machine learning can be solved with relatively high accuracy. There has been a push in research for machine learning models to make accurate but fair decisions. To achieve this goal, there are three main approaches: pre-processing, in-processing, and post-processing methods. 
This work proposes to learn fair low-rank tensor decompositions by regularizing the Canonical Polyadic Decomposition factorization with the kernel Hilbert-Schmidt independence criterion (KHSIC). It is shown, theoretically and empirically, that a small KHSIC between a latent factor and the sensitive features guarantees approximate statistical parity. The proposed algorithm surpasses the state-of-the-art algorithm, FATR (Zhu et al., 2018), in controlling the trade-off between fairness and residual fit on synthetic and real data sets.
\end{abstract}

\section{Introduction}

Tensor factorizations are used in many machine learning applications including link prediction \citep{dunlavy2011temporal}, clustering \citep{shashua2006multi}, and recommendation \citep{kutty2012people}, where they are used to find vector representations (embeddings) of entities. With the widespread use of tensor factorization, we hope that decisions made from using tensor data are accurate but fair. This work focuses on fair Canonical Polyadic Decompositions (CPD) of tensors~\citep{carroll1970analysis}: a rank-$R$ CPD of a three mode tensor $\tX \in \R^{I \times J \times K}$ is defined in terms of factor matrices $\mA \in  \R^{I \times R}$, $\mB \in  \R^{J \times R}$, $\mC \in  \R^{K \times R}$; the corresponding CPD of $\tX$ is  $\llbracket \mA, \mB, \mC \rrbracket = \sum_{i=1}^{R} \mA_{:, i} \circ \mB_{:, i} \circ \mC_{:, i}$, where $\circ$ denotes the outer product and $\mA_{:, i}$, $\mB_{:, i}$, $\mC_{:, i}$ are columns of the factor matrices. The factor matrices are learned to minimize $\|\tX - \llbracket \mA, \mB, \mC \rrbracket \|_\frob^2$, using any of a variety of algorithms~\citep{sidiropoulos2017tensor}.
% A question that is unanswered is how we can guarantee fair and accurate outcomes when using such tensor factorizations. In order to address this problem, we use the framework defined by Dwork et al. (2012). We assume that there are two major entities: a regulator and a vendor. The role of the regulator is to preprocess and construct a fair representation of a data set. The regulator can distribute the preprocessed data set to vendors so that vendors can construct machine learning models which maximize their own utility while guaranteeing some degree of fairness. We take interest in the role of the regulator to answer the following question: How can a fair latent representation of a tensor $\tX$ be constructed?

This work assumes the same setting as~\citet{zhu2018fairness}: one mode of the tensor is associated with a sensitive feature; this mode is called the sensitive mode\footnote{Our approach can be readily extended to dealing with multiple sensitive features and/or modes.}; throughout we fix mode 1, corresponding to $\mA$, to be the sensitive mode. A  CPD is considered fair if the rows of $\mA \in \R^{n \times r}$ are independent of the rows of the sensitive feature matrix $\mS \in \R^{n \times d}$. This implies that the embedding $\mA_{i,:}$ of an entity is independent from its sensitive features, and therefore any function of the entity embeddings will be independent of the sensitive features. In particular, machine learning models built using $\mA$ will satisfy statistical parity.\footnote{See the Appendix for the definition of statistical parity and how it is guaranteed by a fair CPD.}

Despite its common presence in ML applications, and the existence of several approaches for learning fair matrix factorizations (\citet{kamishima2018recommendation}; \citet{kamishima2017considerations}; \citet{kamishima2012enhancement}, to our knowledge there is only one work that directly addresses the problem of learning fair tensor factorizations, \citep{zhu2018fairness}. This work presents the fairness aware tensor recommendation (FATR) algorithm that imposes orthogonality between the columns of $\mA$ and $\mS$:
$$
\mA^\star, \mB^\star, \mC^\star = \argmin_{\mA,\mB,\mC} {\norm{\tX - \llbracket \mA, \mB, \mC \rrbracket}_\frob^2}  + \lambda_o \norm{\mA^\top  \mS}_\frob^2 + \lambda_{\ell_2} \norm{\mA}_\frob^2.$$
The intuition behind FATR is that when $\mA$ and $\mS$ are orthogonal, the entity embeddings are independent from the sensitive variables; this ensures that predictions of any classifier built using the entity embeddings are independent of the sensitive features.

This work is motivated by the observation that this intuition is incorrect.

\begin{theorem}
\label{thm:fatr}
There exists a matrix $\mA \in \R^{n \times r}$ whose rows are drawn from two distinct subpopulations and a matrix $\mS \in \R^{n \times d}$ encoding the subpopulation of the corresponding rows of $\mA$, such that $\mA^T \mS = \bm{0}$ yet there exists a classifier $C$ that can identify which subpopulation a given row of $\mA$ is drawn from with 100\% accuracy. 
\end{theorem}

Theorem~\ref{thm:fatr} demonstrates that the orthogonality constraint of FATR does not guarantee statistical parity in downstream applications. %Indeed, experimental results confirm that FATR does not learn fair tensor factorizations. 
A proof of Theorem~\ref{thm:fatr} is provided in the Appendix.

This work presents a novel algorithm for learning fair CPDs; this approach can be readily extended to higher order tensors and to other tensor factorizations. The nucleus of the approach is the fact that the rows of $\mA$ and $\mS$ define the values random variables $\rva$ and $\rvs$ take over a finite population (the entities). Fairness can therefore be achieved by learning representations of $\rva$ that are independent from $\rvs$, while ensuring low error in approximating $\tX$. It is established, empirically and theoretically, that the proposed algorithm does tradeoff between reconstruction accuracy and fairness.

\section{Canonical Polyadical Decomposition (CPD) with KHSIC regularization}

A classic result of \citet{gretton2005measuring} states that the Kernel Hilbert Space Independence Criterion (KHSIC), given in~\eqref{eqn:KHSCI_def}, of two distributions is zero if and only if they are independent: that is, $\rva$ and $\rvs$ are independent if \emph{all} functions of $\rva$ are uncorrelated with \emph{all} functions of $\rvs$. The quantitative version of this result, given below, further establishes that a small KHSIC between two distributions ensures they are nearly independent. This result is proven in the Appendix. The following theorem holds true for any universal kernel \citep{micchelli2006universal}:

\begin{theorem}
\label{thm:KHSIC}
Let $\rva$ and $\rvs$ be random variables and consider their KHSIC, given by
\begin{equation}
    \label{eqn:KHSCI_def}
    \rho(\rva, \rvs) = \max_{f \in H_1, g \in H_2} \mathbb{E}_{\rva,\rvs}([f(\rva) - \mathbb{E}f(\rva)][g(\rvs) - \mathbb{E}g(\rvs)]),
\end{equation}
where $H_1$ and $H_2$ are universal Reproducing Kernel Hilbert Spaces (RKHSes).
For every set $A$ in the range of $\rva$ and $B$ in the range of $\rvs$,
$$ \lvert \mathbb{P} (\rva \in A \cap \rvs \in B) - \mathbb{P}(\rva \in A) \mathbb{P}(\rvs \in B) \rvert \leq \rho.
$$
Further, if $\rva$ and $\rvs$ are both uniformly distributed over $n$ values, 
%the representer theorem~\citep{scholkopf2001generalized} gives that
\begin{equation}
\label{eqn:KHSIC_form}
\rho(\rva, \rvs) = \frac{1}{n^2}  \langle \tilde{\mK}_\mA, \tilde{\mK}_{\mS} \rangle,
\end{equation}
where $\tilde{\mK}_\mA$ and $\tilde{\mK}_\mS$ are centered $n \times n$ kernel matrices\footnote{See the Appendix for the definition of these kernel matrices and a discussion of kernel selection.} corresponding to $H_1$ and $H_2$. 
\end{theorem}

Thus, a natural formulation for learning fair CPDs is to find
$$\mA^\star, \mB^\star, \mC^\star = \argmin_{\mA, \mB, \mC} \frac{1}{IJK}{\norm{\tX - \llbracket \mA, \mB, \mC \rrbracket}_\frob^2} + \frac{\lambda}{n^2}  \langle \tilde{\mK}_{\mA}, \tilde{\mK}_{\mS} \rangle$$
using block coordinate descent, leading to  Algorithm~\ref{alg:bcd_cpd}. This formulation drives $\rho(\rva, \rvs)$ to zero, thereby increasing the fairness of the tensor factorization, at the expense of raising the reconstruction error of $\tX$.

\begin{algorithm}[H]
\label{alg:bcd_cpd}
\SetKwInput{KwData}{Inputs}
\SetKwInput{KwResult}{Outputs}
\SetAlgoLined
\KwData{Random initializations $\mA_0$, $\mB_0$, $\mC_0$; number of epochs $T$;
 regularization parameter $\lambda$; learning rates $\alpha_0, \alpha_1$}
\KwResult{$\mA$, $\mB$, $\mC$ }
 \For{$j \gets 1,\ldots,T$ }{
        $\mA_{t+1} \gets \mA_{t} - \alpha_0 \nabla _{\mA}\left( {\norm{\tX - \llbracket \mA_t, \mB_t, \mC_t \rrbracket}_\frob^2} + \tfrac{\lambda}{n^2}  \langle \tilde{\mK}_{\mA_t}, \tilde{\mK}_{\mS} \rangle\right)$  \\
        $\mB_{t+1} \gets \mB_{t} - \alpha_1 \nabla _{\mB}{\norm{\tX - \llbracket \mA_{t+1}, \mB_t, \mC_{t}\rrbracket}_\frob^2}$ \\
        $\mC_{t+1} \gets \mC_{t} - \alpha_1 \nabla_{\mC} {\norm{\tX - \llbracket \mA_{t+1}, \mB_{t+1}, \mC_t \rrbracket}_\frob^2 }$\\
 }
 \caption{KHSIC-regularized Block Coordinate Descent (KHSIC-BCD)}
\end{algorithm}

\section{Experiments}
This section documents experiments conducted on synthetic and real data sets to illustrate the trade-offs between the residual fit and fairness for the KHSIC-BCD algorithm. Unregularized BCD, FATR, and HSIC-BCD (in which HSIC\footnote{The Hilbert Space Independence Criterion, or HSIC, is the squared Frobenius norm of the correlation matrix of $\rva$ and $\rvs$, $\|\tilde{\mA}^T \tilde{\mS}\|_\frob^2$. This is very similar to the orthogonality constraint of FATR, and if zero, implies that $\rva$ and $\rvs$ are uncorrelated.} is used as a regularizer instead of KHSIC) are used as baselines for comparison.  As the experiments use only binary sensitive features, the fairness of tensor factorizations is quantified by the ability of a classifier to identify the value of a sensitive attribute, given the corresponding rows of $\mA$:
\begin{align*}
    \text{unfairness} = \left(\frac{\text{\# correctly identified attributes}}{\text{\# total number of attributes}} - 0.5\right)
\end{align*}
Lower unfairness corresponds to higher approximate statistical parity.
Due to space considerations, the descriptions of the other metrics used to evaluate the methods, the experimental setup, and descriptions of the data set have been deferred to the Appendix.

\subsection{Synthetic Experiments}
We explored the effects of minimizing the orthogonality constraint from FATR and the KHSIC from KHSIC-BCD upon achieving fairness; this is accomplished in both cases by increasing the regularization parameters. Figure 1a shows that as we minimize the orthogonality constraint of FATR, there is no improvement in fairness. In contrast, we see in Figure 1b that the unfairness metric decreases as $\hat{\rho}(\mA,\mS)$ goes to zero in the proposed KHSIC-BCD algorithm, which implies the classifier cannot identify the subpopulation associated with a given row of $\mA$. This experiment was conducted on synthetic data described in the Appendix.

\begin{figure}%
    \centering
    \subfloat[ Unfairness vs FATR orthogonality. Unfairness remains unchanged even as the orthogonality constraint goes to zero. ]{{\includegraphics[width=4.8cm]{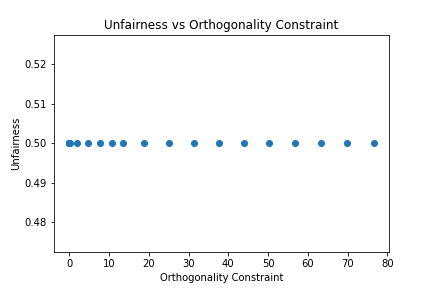} }}%
    \qquad
    \subfloat[ Unfairness vs KHSIC. As $\hat{\rho}(\mA,\mS)$ goes to zero, the classifier fails to distinguish between the two subpopulations.] {{\includegraphics[width=4.8cm]{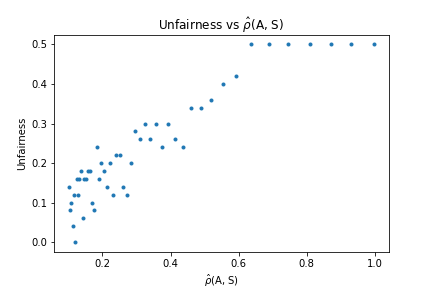} }}%
    \caption{Experiments comparing FATR and KHSIC-BCD}%
    \label{fig:fairnessvsmetric}%
\end{figure}

We also looked at the trade-off between residual fit and fairness on the synthetic dataset. Figure 2a shows the Pareto frontier between unfairness and residual fit. The plot shows that when the rows of the latent factor matrix correspond to two completely different distributions, we must sacrifice a considerable amount of approximation accuracy to guarantee fairness. KHSIC-BCD outperforms HSIC-BCD by learning latent factorizations with higher fit quality given a fixed degree of unfairness. 

Lastly, we compared the residual error and fairness across all four methods on the synthetic dataset, averaged over five runs. Figure 2b shows that FATR provides a higher residual error with no improvement in fairness. KHSIC-BCD has lower residual errors and achieves more fairness than HSIC-BCD.

\subsection{Contraceptive Data Set}
Using a real data set, we constructed the Pareto frontier between unfairness and residual error; see Figure 3a. This plot shows that the unfairness metric can be minimized while sacrificing a small amount of residual fit quality by utilizing HSIC and KHSIC regularizations. 

We also compare the residual error and fairness across all four methods. Figure 3b shows that HSIC-BCD and KHSIC-BCD can be used to minimize the unfairness metric while obtaining a factorization whose residual error is comparable to that of BCD. Using FATR to minimize the orthogonality constraint increases the residual error and gives no more fairness than BCD. 

\begin{figure}%
    \centering
    \subfloat[ Pareto frontier between fairness and relative residual error. ]{{\includegraphics[width=4.8cm]{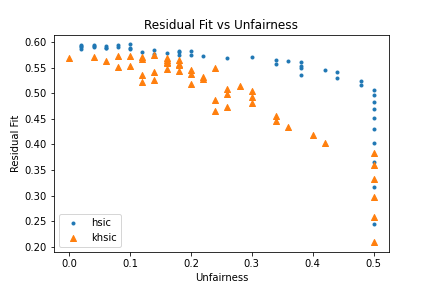} }}%
    \qquad
    \subfloat[ The relative residual errors and fairness of BCD, FATR, HSIC-BCD, and KHSIC-BCD.]{{\includegraphics[width=4.8cm]{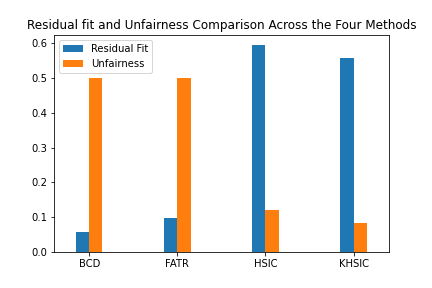} }}%
    \caption{Experiments on Synthetic Data Set}%
    \label{fig:synthetic}%
\end{figure}

\begin{figure}%
    \centering
    \subfloat[ Pareto frontier between fairness and relative residual error. ]{{\includegraphics[width=4.8cm]{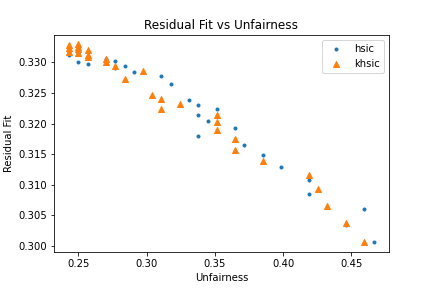} }}%
    \qquad
    \subfloat[ The relative residual errors and fairness of BCD, FATR, HSIC-BCD, and KHSIC-BCD.]{{\includegraphics[width=4.8cm]{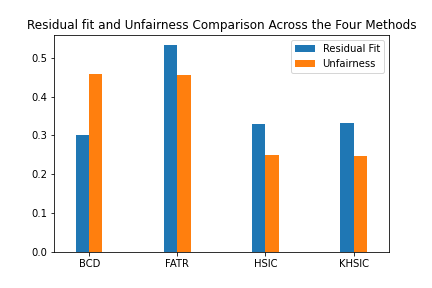} }}%
    \caption{Experiments on Contraceptive Data Set}%
    \label{fig:contraceptive}%
\end{figure}

\section{Conclusion}
An algorithm was presented that facilitates the learning of fair CP decompositions of tensors by trading off between fairness and residual fit. Experimental and theoretical evidence show that the proposed KHSIC-BCD algorithm significantly outperforms FATR in enforcing fairness, and the extent to which a small KHSIC guarantees approximate independence was quantified. Future work includes extending this framework to fair tensor completion and fair exponential family tensor completion, to facilitate working with discrete-valued datasets.

\newpage
%\nocite{*}
\bibliographystyle{iclr2021_conference}
%\bibliography{iclr2021_conference}

\input{paper.bbl}
\newpage
\appendix
\section{Appendix}

\subsection{Notation}
Bold roman uppercase letters (e.g. $\tX$) denote tensors, bold uppercase letters denote matrices (e.g. $\mX$), and bold lowercase letters (e.g. $\va$) denote vectors. We denote the $i^{\text{th}}$ column and $j^{\text{th}}$ row of a matrix by $\mX_{:, i}$ and  $\mX_{j, :}$, respectively.  The Frobenius norm is denoted by $\norm{\cdot}_\frob$, and the matrix inner product is defined as $\langle \mA, \mB \rangle = \mathrm{Tr}(\mA \mB^T)$.

\subsection{Statistical Parity}
The goal of a fair CPD is to ensure statistical parity in classifiers built using the entity embeddings derived from the tensor factorization. Let a classifer $C$ take an entity embedding $\rva$ as input, and let $\rvs$ be the corresponding sensitive feature; $C$ is said to have statistical parity if the classifier's output is independent of the value of the sensitive variable:
\[
\mathbb{P}(C(\rva) = 1 \,|\, \rvs) = \mathbb{P}(C(\rva) = 1),
\]
and has approximate statistical parity up to bias $\varepsilon$ if 
\[
|\mathbb{P}(C(\rva) = 1 \,|\, \rvs) - \mathbb{P}(C(\rva) = 1)| < \varepsilon.
\]
Functions of independent random variables are independent, so  $\rva \indep \rvs$ implies $C(\rva) \indep \rvs$. Thus to ensure that \emph{any} classifer $C$ has statistical parity, it suffices to ensure that the distribution of the rows of $\mA$ is independent of the distribution of the rows of $\mS$.

Similarly, approximate statistical parity is achievable using a CPD where the distribution of the rows of $\mA$ is approximately independent of the distribution of the rows of $\mS$ in the sense of Theorem~\ref{thm:KHSIC}. As an example, assume\footnote{Such a $\gamma$ exists if, for example, $\rvs$ is atomic, as in applications where $\rvs$ is a categorical variable denoting group membership.} $\min_{S} \mathbb{P}(\rvs \in S) = \gamma >0$. Then for any set $S$,
\begin{align*}
    | \mathbb{P}( C(\rva) = 1\,|\, \rvs \in S) - \mathbb{P}(C(\rva) = 1)| & = 
    \left| \frac{\mathbb{P}( C(\rva) = 1 \cap \rvs \in S)}{\mathbb{P}(\rvs \in S)} - \frac{\mathbb{P}( C(\rva) = 1) \mathbb{P}(\rvs \in S)}{\mathbb{P}(\rvs \in S)} \right| \\
    & \leq \gamma^{-1} |\mathbb{P}( C(\rva) = 1 \cap \rvs \in S) - \mathbb{P}( C(\rva) = 1) \mathbb{P}(\rvs \in S) | \\
    & = \gamma^{-1} |\mathbb{P}( \rva \in C^{-1}(1) \cap \rvs \in S) - \mathbb{P}( \rva \in C^{-1}(1)) \mathbb{P}(\rvs \in S)| \\
    & \leq \gamma^{-1}\rho.
\end{align*}

% You may include other additional sections here.
\subsection{Centered Kernel Matrix and Kernel Selection}
Given a kernel $k_1$ corresponding to the RKHS $H_1$, define the kernel matrix $\mK_\mA$ satisfying $(\mK_\mA)_{ij} = k_1(\mA_{i,:}, \mA_{j,:})$, and define the centered kernel matrix $\tilde{\mK}_\mA = \mH\mK_\mA \mH$, where $\mH = \mI - \frac{1}{n} \bf{1} \bf{1}^\top$. Let $k_2$ and $\tilde{\mK}_{\mS}$ be defined similarly. 

To satisfy Theorem 2, any universal kernel can be chosen. Throughout the paper, we utilize the radial basis kernel $k(\vx,\vy) = \exp{(-\gamma \norm{\vx-\vy}_2^2})$, as it is a widely used universal kernel.
\subsection{Proof for Thereom 1}
\begin{proof}
A latent factor matrix satisfies FATR's orthogonality constraint if the columns of the factor matrix $\mA \in \R^{n \times r}$ are orthogonal to the columns of the sensitive matrix $S \in \R^{n\times d}$.

Define the following matrices:
\[
    A = \begin{bmatrix}
        1 & 0 & \dots & 0 \\
        0 & 0 & \dots & 0 \\
        0 & 0 & \dots & 0 \\
        -1 & 0 & \dots & 0 \\
        -1 & 0 & \dots & 0 \\
        1 & 0 & \dots & 0 \\
        \end{bmatrix} \quad
    S = \begin{bmatrix}
        1 & 0 \\
        0 & 1 \\
        0 & 1 \\
        1 & 0 \\
        1 & 0 \\
        1 & 0 \\
        \end{bmatrix}
\]

In this example, we assume that there are two subpopulations where one is labeled by 1 and the other is labeled by 0. Then the columns of $\mS$ simply contain the subpopulation labels for each row of the latent matrix $\mA$. Clearly, the columns of $\mA$ are orthogonal to every column of $\mS$ so FATR's orthogonality constraint is met. \\
Define a classifer C where
\begin{align*}
    C(\va_i) = \norm{\va_1}  
\end{align*}

Applying this classifier on every row of $\mA$ successfully extracts the the sensitive labeling of the rows of $\mA$ with 100\% accuracy. 
\end{proof}

This is an example where FATR can learn a sensitive latent matrix $\mA$ that is not independent of the sensitive features; therefore, we use this as motivation to find methods in learning fair tensor factorization. Note that FATR is concerned with learning a latent representation of a tensor with missing data; however, FATR fails to learn a fair representation for a tensor that is fully observed.

\subsection{Proof of Theorem 2}
\begin{proof}
Fix arbitrary sets A in the range of $\va$ and B in the range of $\vs$, and an arbitrary $\epsilon \in (0,1]$. Because $H_1$ and $H_2$ are c-universal RKHSes (Sriperumbudur et al. 2011), there exists $f_\epsilon \in H_1$ and $g_\epsilon \in H_2$ such that
$\norm{f_\epsilon- \mathbbm{1}_A}_\infty \leq \epsilon$ and $\norm{g_\epsilon- \mathbbm{1}_B}_\infty \leq \epsilon$,
where $\mathbbm{1}_A$ and $\mathbbm{1}_B$ are the 0-1 indicator functions for A and B.

The KHSIC bounds the correlation of $f_\epsilon(\va)$ and $g_\epsilon(\vs)$ since
\begin{align}
    \rho &\geq \lvert \mathbb{E}([f_\epsilon(\va) - \mathbb{E}f_\epsilon(\va)][g_\epsilon(\vs) - \mathbb{E}g_\epsilon(\vs)]) \rvert \nonumber \\
    &=  \lvert \mathbb{E}[f_\epsilon(\va) g_\epsilon(\vs)] - [\mathbb{E} f_\epsilon(\va)][\mathbb{E} g_\epsilon(\vs)] \rvert \label{eqn:corrbound}
\end{align}

Notice that, by Jensen's inequality
\begin{equation}
   \label{eqn:probA}
    \lvert \mathbb{E} f_\epsilon(\va ) - \mathbb{P} (\va \in A) \rvert = \lvert \mathbb{E} f_\epsilon(\va) - \mathbb{E} \mathbbm{1}_A(\va) \rvert \leq \mathbb{E} \lvert f_\epsilon(\va) - \mathbbm{1}_A(\va) \rvert \leq \epsilon 
\end{equation}
and similarly, 
\begin{equation}
    \mathbb{E} \lvert g_\epsilon(\vs ) - \mathbb{P} (\vs \in B) \rvert \leq \mathbb{E} \lvert g_\epsilon(\vs) - \mathbbm{1}_B(\vs) \rvert \leq \epsilon. \label{eqn:probB}
\end{equation}

By Jensen's inequality and a judicious application of the triangle inequality,
\begin{align}
    \lvert \mathbb{E}[f_\epsilon(\va) g_\epsilon(\vs)] - \mathbb{P}(\va \in A \cap \vs \in B)] \rvert \nonumber &\leq \mathbb{E} \lvert f_\epsilon(\va) g_\epsilon(\vs) - \mathbbm{1}_A(\va)  \mathbbm{1}_B(\vs) \rvert  \nonumber  \\
    &\mkern-200mu \leq \mathbb{E} \lvert[f_\epsilon(\va) - \mathbbm{1}_A(\va)]g_\epsilon(\vs)  \rvert  \nonumber + \mathbb{E} \lvert[g_\epsilon(\vs) - \mathbbm{1}_B(\vs)]\mathbbm{1}_A(\va) \rvert   \nonumber \\
    &\mkern-200mu \leq 2\epsilon (1+\epsilon), \label{eqn:probABdep}
\end{align}
where the final inequality is justified by estimates~(\ref{eqn:probA}) and~(\ref{eqn:probB}).

Estimates~(\ref{eqn:probA}) and (\ref{eqn:probB}) also imply that 
\begin{align}
     \lvert \mathbb{P}(\va \in A) \mathbb{P}(\vs \in B) -  [\mathbb{E} f_\epsilon(\va)][\mathbb{E} g_\epsilon(\vs)]\rvert & \nonumber \\
     & \mkern-200mu \leq \lvert \mathbb{P}(\va \in A) [\mathbb{P}(\vs \in B) -  \mathbb{E} g_\epsilon(\vs)]\rvert  \nonumber + \lvert  \mathbb{E} g_\epsilon(\vs) [\mathbb{P}(\va \in A) -  \mathbb{E} f_\epsilon(\va)]\rvert  \nonumber\\
     &\mkern-200mu \leq 2\epsilon(1+\epsilon). \label{eqn:probABindep}
\end{align}

Using estimates~(\ref{eqn:probABdep}),~(\ref{eqn:probABindep}), and~(\ref{eqn:corrbound}), we find that 
\begin{align*}
    \lvert \mathbb{P} (\va \in A \cap \vs \in B) - \mathbb{P}(\va \in A) \mathbb{P}(\vs \in B) \rvert & \\
    &\mkern-264mu \leq \lvert \mathbb{P} (\va \in A \cap \vs \in B) -  \mathbb{E}[f_\epsilon(\va) g_\epsilon(\vs)] \rvert + \lvert \mathbb{P}(\va \in A) \mathbb{P}(\vs \in B) - [\mathbb{E} f_\epsilon(\va )][\mathbb{E} g_\epsilon(\vs)] \rvert \\
    & \mkern-240mu + \lvert \mathbb{E}[f_\epsilon(\va) g_\epsilon(\vs)] - [\mathbb{E} f_\epsilon(\va)][\mathbb{E} g_\epsilon(\vs)]  \rvert \\
    &\mkern-264mu \leq 4\epsilon(1+\epsilon) + \rho. 
\end{align*}

Since $\epsilon$ can be taken arbitrarily close to zero, we conclude that as claimed,
\begin{align*}
    \lvert \mathbb{P} (\va \in A \cap \vs \in B) - \mathbb{P}(\va \in A) \mathbb{P}(\vs \in B) \rvert \leq \rho.
\end{align*}

The fact that $\rho$ is given by~\eqref{eqn:KHSIC_form} is a consequence of the representer theorem~\citep{scholkopf2001generalized}.
\end{proof}

\subsection{Experimental metrics}

To measure how well the latent factorization approximated the original tensor, we use the normalized frobenius norm error
\begin{align*}
    \text{residual fit} = \frac{\norm{\tX - [[\mA, \mB, \mC]]}_\frob}{\norm{\tX}_\frob}
\end{align*}

To measure FATR's independence criterion, we use their definition of the orthogonality constraint
\begin{align*}
    \text{orthogonality constraint} = \norm{\mA^\top \mS}_\frob
\end{align*}
% As a practical measure of independence between $\mA$ and $\mS$, we measured if it was possible for a classifier to correctly identify the sensitive variables from the rows of $\mA$. Because we only considered binary sensitive features, we used the following metric:
As stated in the earlier experimental section, we measured if it was possible for a classifier to correctly identify the sensitive variables from the rows of $\mA$.
\begin{align*}
    \text{unfairness} = \left(\frac{\text{\# correct predictions}}{\text{\# total predictions}} - 0.5\right)
\end{align*}
where we predict the binary label of the rows of $\mA$ on the test data. If the rows of $\mA$ are truly independent from $\mS$ and the training data was balanced, the classifier would randomly assign a label to every row. Therefore, the unfairness metric achieves a minimum value of 0 when the rows of $\mA$ are independent from those of $\mS$. 

For an unbalanced training data set, we can also expect that the classifier simply guesses the majority label to maximize its classification accuracy. This still corresponds to a situation where a classifier cannot identify between the two sub populations. Therefore, the unfairness metric could in that case converge to
\begin{align*}
    \text{unfairness} &= \left(\frac{\text{\# majority labels in test set}}{\text{\# total predictions}} - 0.5\right)
\end{align*}

%We want to emphasize that \emph{lower} values of the unfairness metric corresponds to a higher degree of independence between the rows of the latent factor matrix $\mA$ and $\mS$

A two-layer neural network was used as the classifier used to measure unfairness:
\begin{align*}
    N(\mA) = \text{softmax}(\text{relu}(\mA\mW_1 + \mB) \mW_2);
\end{align*}
here $\mW_1 \in \R^{R \times 1500}$ and $\mW_2 \in \R^{1500 \times 2}$. 

For the synthetic data set, the neural network was trained by randomly sampling 75\% of the rows of $\mA$ as training data and running 100 epochs of SGD with a learning rate of 0.003. The accuracy of the classifier was computed over the other 25\% as test data. For the contraceptive data set, we randomly split the data set into 90\% training data and 10\% test data, and train using SGD with the same parameter settings.

For interpretability, we report the normalized KHSIC:
\begin{align*}
    \hat{\rho}(\mA, \mS) &= \frac{\langle \tilde{\mK}_{\mA}, \tilde{\mK}_{\mS} \rangle}{\norm{\tilde{\mK}_{\mA}}_\frob\norm{\tilde{\mK}_{\mS}}_\frob}.
    % HSIC_{cos}(\mA, \mS) &= \frac{\langle \tilde{\mA}\tilde{\mA}^\top, \tilde{\mS} \tilde{\mS}^\top\rangle}{\norm{\tilde{\mA}\tilde{\mA}^\top}_\frob\norm{\tilde{\mS} \tilde{\mS}^\top}_\frob}
\end{align*}
This is the centered kernel alignment defined in \citet{cortes2012algorithms}, which measures the cosine of the angle between the centered kernel matrices.

\subsection{Implementation Details}
All algorithms were implemented using Pytorch. For simplicity, Algorithm~\ref{alg:bcd_cpd} takes a single gradient step for the factor matrices in every epoch. We found that taking multiple gradient steps per factor matrix provided better results. In our experiments, 200 gradient steps were utilized to update a factor matrix before moving on to the succeeding factor matrix. 

\subsection{Details of the Synthetic Data Set Experiment}

Three latent factor matrices $\mA_x \in \R^{200 \times 10}$, $\mB_x \in \R^{100 \times 10}$, and $\mC_x \in \R^{100 \times 10}$ were generated to construct a tensor $\tX \in \R^{200 \times 100 \times 100}$. Each row of $\mA_x$ was sampled from two distributions: $\mathcal{N}( \bm{\mu}, \bm{\Sigma})$ with $\bm{\mu}= [1;2;\ldots;10]^T$ and $\bm{\Sigma} = \mathrm{diag}([1; 0.9;\ldots;0.1]^T)$, and a uniform distribution U(0,1). 50\% of the rows were sampled from the normal distribution while the other 50\% of the rows were sampled from the uniform distribution. The entries of $\mB_x$ and $\mC_x$ were generated from a uniform distribution U(0,1). To learn a low rank approximation of this tensor, we take $\mA_0 \in \R^{200 \times 8}, \mB_0 \in \R^{100 \times 8}$, and $\mC_0 \in \R^{100 \times 8}$ where each entry comes from the uniform distribution U(0,1).  The kernel chosen for KHSIC-BCD was the radial basis function (RBF) kernel with $\gamma = 1$.

The learning rate for updating $\mA$ was $\alpha_0 = 3$ for non-regularized BCD and FATR and $\alpha_0 = 0.5$ for HSIC-BCD and KHSIC-BCD. The learning rate for updating $\mB$ and $\mC$ was $\alpha_1 = 1$ across all methods. The $\ell_2$ regularization parameter for FATR was chosen as $\alpha_{l2} =  1$. To generate Figure 1a and Figure 1b, 20 points were sampled linearly between [0, 0.1] for the orthogonal regularization parameter $\alpha_o$ and 100 points were sampled linearly from [0, .04] for the KHSIC regularization parameter respectively.  To generate Figure 2a, we linearly sampled 50 regularization parameters from [0, 0.4] for KHSIC-BCD and [0, 0.01] for HSIC-BCD. 
 
 To generate Figure 2b, the regularization parameters controlling fairness were $\alpha_o = 0.1$, $\lambda_h = 0.01$, $\lambda_k = 0.04$ for FATR, HSIC-BCD, and KHSIC-BCD respectively. These regularization parameters were chosen to put more importance on the fairness regularizer. The FATR regularization parameter was chosen such that $\norm{\mA^\top \mS}_\frob < 10^{-6}$ and the HSIC-BCD and KHSIC-BCD regularization parameters are the largest value in the range we tested in generating the Pareto frontier.
 
 \subsection{Details of the Contraceptive Data Set Experiment}
 The contraceptive method data set~\citep{Dua2019uci} is used to predict an Indonesian woman's choice of contraception given their age, education, etc. The data set consists of 1473 samples with 9 features. We chose to filter the data set of a woman's occupation status. 74.9\% of women in the data set were working (1 label) and the other 25.1\% were not working (0 label). To learn a low rank approximation of this matrix, we generate factor matrices $\mA_0 \in \R^{1473 \times 6}$ and $\mB_0 \in \R^{9 \times 6}$, where each entry comes from the uniform distribution U(0,1). The learning rate for updating $\mA$ and  $\mB$ was $\alpha_0 = 100$ and $\alpha_1 = 10$ respectively across all methods.  The kernel chosen for KHSIC-BCD was the radial basis kernel with $\gamma = 1$.

To generate Figure 3a, we linearly sampled 15 regularization parameters from [0, 0.66] and 15 regularization parameters from [0.66, 4.8] for both HSIC-BCD and KHSIC-BCD.

To generate Figure 3b, the orthogonality regularization parameter and $\ell_2$ regularization parameter for FATR was chosen as $\alpha_o = 0.00002$ and $\alpha_{l2} =  1$. The orthogonality regularization parameter was picked such that 
$\norm{\mA^\top \mS}_\frob < 0.1$. The regularization parameters for KHSIC-BCD and HSIC-BCD were $\lambda = 4.8$; this was the largest value tested in generating the Pareto frontier graph. Once again, the largest regularization parameters were chosen so that we can compare the effects of trading off residual fit quality and fairness.

\section{Related Works}
The literature most closely related to this work focuses on learning fair latent representations (\citet{zhu2018fairness}; \citet{kamishima2012enhancement}; \citet{kamishima2017considerations}; \citet{kamishima2018recommendation}). Zhu et al.~used orthogonality constraints to encourage fairness. Kamishima et al.~learned fair latent representations for recommender systems. In order to have the latent features of an entity to be independent from its sensitive features, Kamishima et al.'s work has proposed mean matching between subpopulations of the rows of the latent matrix $\mA$, distribution matching with Bhattacharyya distance, and mutual information metrics to decrease the dependence between the latent features and the sensitive features~\citep{kamishima2012enhancement, kamishima2017considerations, kamishima2018recommendation}.

\citet{perez2017fair} similarly proposes to learn fair latent representations using the KHSIC criterion. In contrast to our proposed algorithm, the approach of~\citet{perez2017fair} applies only to matrices and is formulated as a generalized eigenvalue problem where the tradeoff between approximation error and fairness cannot be explicitly controlled. 

\end{document}

%% file: math_commands.tex
\usepackage{amsmath,amsfonts,bm}

\def\eqref#1{equation~\ref{#1}}
\def\1{\bm{1}}

\def\rva{{\mathbf{a}}}

\def\rvs{{\mathbf{s}}}

\def\va{{\bm{a}}}

\def\vs{{\bm{s}}}

\def\vx{{\bm{x}}}
\def\vy{{\bm{y}}}

\def\mA{{\bm{A}}}
\def\mB{{\bm{B}}}
\def\mC{{\bm{C}}}

\def\mH{{\bm{H}}}
\def\mI{{\bm{I}}}

\def\mK{{\bm{K}}}

\def\mS{{\bm{S}}}

\def\mW{{\bm{W}}}
\def\mX{{\bm{X}}}

\DeclareMathAlphabet{\mathsfit}{\encodingdefault}{\sfdefault}{m}{sl}
\SetMathAlphabet{\mathsfit}{bold}{\encodingdefault}{\sfdefault}{bx}{n}
\newcommand{\tens}[1]{\bm{\mathsfit{#1}}}

\def\tX{{\tens{X}}}

\newcommand{\R}{\mathbb{R}}

\newcommand{\frob}{\mathrm{F}}

\DeclareMathOperator*{\argmin}{arg\,min}

\newcommand{\indep}{\perp \!\!\! \perp}